\documentclass{article}

\usepackage{arxiv}

\usepackage[utf8]{inputenc} 
\usepackage[T1]{fontenc}    
\usepackage{hyperref}       
\usepackage{url}            
\usepackage{booktabs}       
\usepackage{amsfonts}       
\usepackage{nicefrac}       
\usepackage{microtype}      
\usepackage{lipsum}
\usepackage{graphicx}
\graphicspath{ {./images/} }
\usepackage{amsmath}

\title{A Novel Method for Improving Accuracy in Neural Network by
Reinstating Traditional Back Propagation Technique}

\author{
 Gokulprasath R \\
  Department of Computer Science and Engineering\\
  Sri Manakula Vinayagar Engineering College\\
  Madagadipet, Puducherry- 605 107.  \\
  \texttt{gokulprasathaids@smvec.ac.in} \\
}

\begin{document}
\maketitle
\begin{abstract}
Deep learning has revolutionized industries like computer vision, natural language processing, and speech recognition. However, back propagation, the main method for training deep neural networks, faces challenges like computational overhead and vanishing gradients. In this paper, we propose a novel instant parameter update methodology that eliminates the need for computing gradients at each layer. Our approach accelerates learning, avoids the vanishing gradient problem, and outperforms state-of-the-art methods on benchmark data sets. This research presents a promising direction for efficient and effective deep neural network training.
\end{abstract}


\section{Introduction}
Deep learning has revolutionized the field of artificial intelligence by enabling machines to learn complex patterns and perform tasks that were previously deemed impossible.  However, training deep neural networks is a challenging and computationally expensive task that requires optimizing millions or even billions of parameters . The back propagation algorithm has been the go-to method for training \cite{talukder2023efficient} deep neural networks for decades, but it suffers from some limitations, such as slow convergence and the vanishing gradient problem.
To overcome these limitations, several alternative training methods have been proposed, such as Standard Back propagation and Direct Feedback Alignment. The core idea of this approach is to update the weights and biases in each layer of a neural network using the local error at that layer, rather than back propagating the error from the output layer to the input layer.\cite{liu2021activated} By doing so, the training process can be accelerated and the model's accuracy can be improved.

In this paper, we propose a novel approach for layer-wise error calculation and parameter update in neural networks and evaluate its performance on a benchmark data set. We compare our approach to other existing methods, such as back propagation, and demonstrate its effectiveness in terms of convergence speed and accuracy. The rest of the paper is organized as follows.\cite{shulman2023optimization} In Section 2, we provide a brief review of the related work on layer-wise error calculation and parameter update in neural networks. In Section 3, we present our proposed approach and its mathematical formulation. In Section 4, we describe the experimental setup and present the results of our evaluation. Finally, in Section 5, we conclude the paper and discuss potential avenues for future research. 

\section{Literature Review}

Back propagation and Direct Feedback Alignment (DFA) are two widely used methods for training artificial neural networks. Both methods aim to adjust the weights of the network to minimize the difference between the predicted output and the actual output. In this literature review, we will compare and contrast these two methods, highlighting their strengths and weaknesses.\cite{zeiler2012adadelta} Backpropagation is a commonly used algorithm for training neural networks. The basic idea behind
back propagation is to calculate the error at the output layer and propagate it backward through the network, adjusting the weights of each neuron in the network to minimize the error. The back propagation algorithm is computationally efficient and can be used to train deep neural networks with many layers. However, back propagation has several limitations. One of the main limitations is that it requires the computation of the derivative of the activation function at each layer, which can be computationally expensive. Additionally, back propagation is prone to getting stuck in local minima, which can lead to sub optimal solutions. 

Direct Feedback Alignment (DFA) is a newer method for training neural networks that does not rely on back propagation. Instead of using the gradient of the loss function to update the weights, DFA uses a fixed random matrix to propagate the error from the output layer back to the hidden layers.\cite{kingma2017adam} This random matrix is learned during the training process and is used to update the weights of the hidden layers. DFA has several advantages over back propagation. One of the main advantages is that it is less computationally expensive, as it does not require the computation of the derivative of the activation function at each layer. Additionally, DFA is less prone to getting stuck in local minima than back propagation. Several studies have compared the performance of back propagation and DFA on various tasks. A study by \cite{nøkland2016direct} found that DFA performed as well as back propagation on several benchmark data sets, including MNIST and CIFAR-10

\section{Proposed Approach}

The approach involves four main steps: forward pass, error calculation, parameter update, and repetition. The forward pass computes the outputs of each layer using the current weights and biases and passing them into the activation function. The error calculation step computes the error at each layer using a layer-wise loss function that takes into account the local deviation between the predicted and target values of that layer. The parameter update step updates the weights and biases of each layer using the calculated error and a layer-wise learning rate that controls the magnitude of the update. Finally, the repetition step repeats the first three steps for multiple epochs or until convergence.

\begin{figure}[h!]
    \centering
    \includegraphics[width=0.75\linewidth]{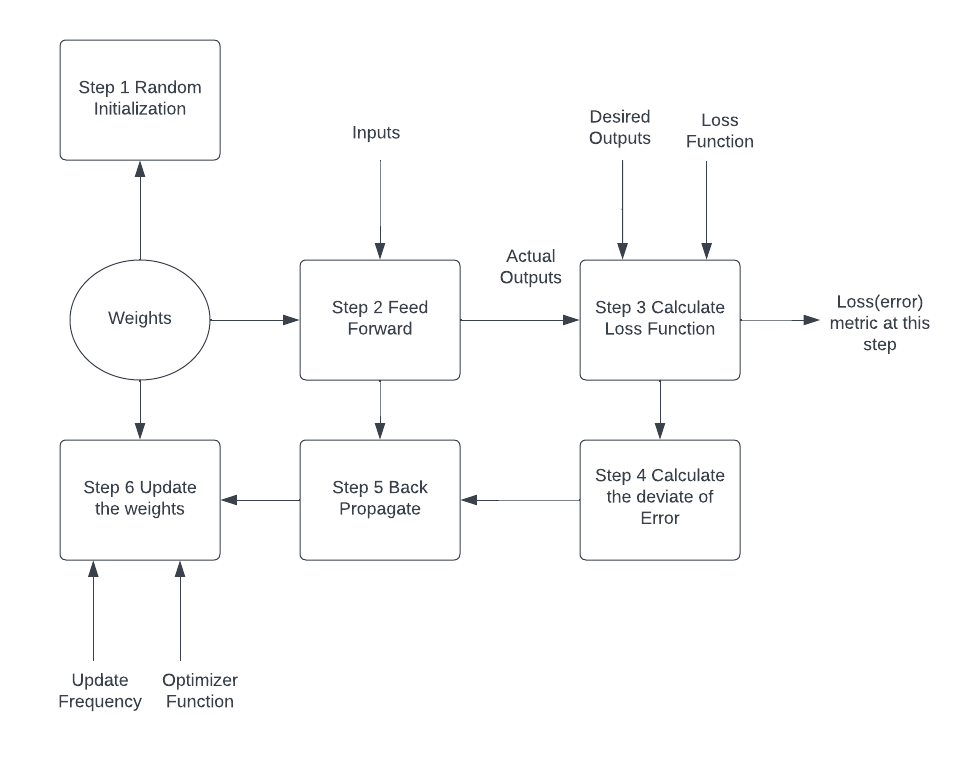}
    \caption{Neural Network Life cycle}
    \label{fig:label1}
\end{figure}

The proposed approach differs from the traditional backpropagation algorithm, which calculates the error at the output layer and backpropagates it to update the parameters of all the layers. The layer wise approach allows for more localized and efficient updates that can potentially accelerate the training process and avoid the vanishing and exploding gradient problem. However, it requires careful tuning of the layer-wise loss function and learning rate, as well as a suitable initialization of the parameters.

\subsection{Forward Pass}
Compute the activations of each layer using the current weights and biases, starting from the
input layer and propagating forward to the output layer. 

\textbf{Algorithm:}
\begin{enumerate}
  \item Initialize the input activation as the input data: $h_0 = x$
  \item For each layer $I$ in the neural network:
  \begin{enumerate}
    \item Calculate the pre-activation value: $z_I = W_{I-1}h_{I-1} + b_I$
    \item Calculate the activation value using an activation function: $h_I = f(z_I)$, where $f$ is a non-linear function such as ReLU, sigmoid, or tanh.
  \end{enumerate}
  \item Output the activations for each layer: $h_I$
\end{enumerate}
\begin{figure}[h!]
    \centering
    \includegraphics[width=0.75\linewidth]{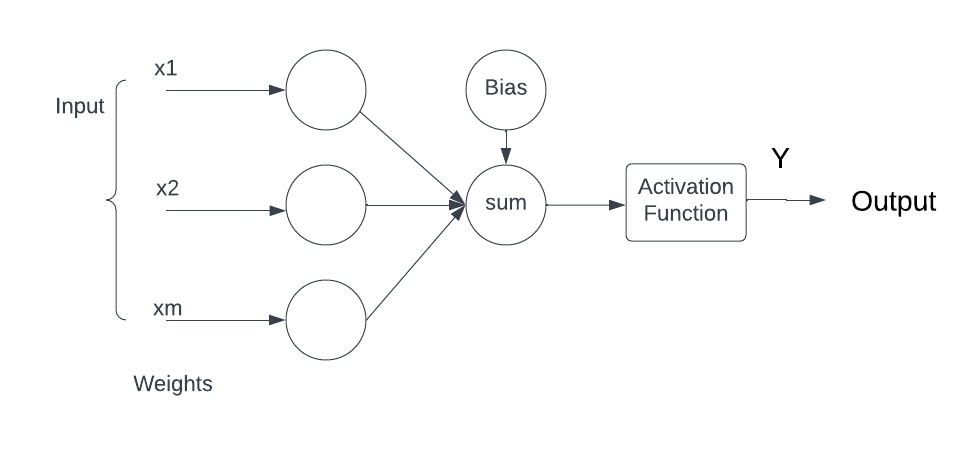}
    \caption{Forward Propagation}
    \label{fig:label2}
\end{figure}
The pre-activation value $z_l$ is the weighted sum of the activations from the previous layer $h_{l-1}$, plus the bias term $b_I$. The activation function $f$ transforms the pre-activation value $z_l$ into the activation value $h_I$, which is then passed on to the next layer. The choice of activation function depends on the specific task and architecture of the neural network, but common choices include ReLU, sigmoid, and tanh. The formulas for calculating the pre-activation value and activation value are:
\begin{align*}
  \text{Pre-activation value: } z_I &= W_Ih_{I-1} + b_I \\
  \text{Activation value: } h_I &= f(z_I)
\end{align*}
where $W_I$ is the weight matrix for layer $I$, $b_I$ is the bias vector for layer $I$, $h_{I-1}$ is the activation vector from the previous layer, and $f$ is the activation function.

\subsection{Error Calculation}
Compute the error at each layer using a layer-wise loss function that takes into account the local deviation between the predicted and target values of that layer. The layer-wise loss function can be
defined based on the specific task and architecture of the neural network, but it should capture the local errors that are relevant for updating the parameters of that layer.

\textbf{Algorithm:}
\begin{enumerate}
  \item Initialize an empty list to store the error of each layer.
  \item For each layer $L$ in the neural network, do the following:
  \begin{enumerate}
    \item Compute the predicted values of the layer $L$ using its activations and the current weights and biases: $Z_L = W_L \cdot A_{L-1} + b_L = activation\_function(Z_L)$
    \item Compute the target values of layer $L$: $T_L = targets$ of layer $L$
    \item Compute the layer-wise loss function for layer $L$: $loss_L = layerwise\_loss\_function(A_L, T_L)$
    \item Compute the error of layer $L$: \\
    $\delta_L = derivative (i.e.) Z_L \cdot (1 - Z_L)$. \\
    $\delta_{errorL} = \delta_L \cdot transpose(W_{L+1})$
    \item Add $\delta_{errorL}$ to the list of errors.
  \end{enumerate}
  \item Return the list of errors.
\end{enumerate}

In the above algorithm, $W_L$ and $b_L$ represent the weights and biases of layer $L$, $A_{L-1}$ and $A_L$ represent the activations of the previous layer and the current layer, respectively. $Z_L$ is the weighted input to layer $L$, $T_L$ is the target values for layer $L$, and $\delta_L$ is the error signal for layer $L$. The activation function and its derivative are denoted as $activation\_function$ and $derivative$, respectively, and the layer-wise loss function and its derivative are denoted as $layerwise\_loss\_function$ and $derivative$, respectively. The $transpose(W_{L+1})$ term in step 2d represents the transpose of the weights connecting layer $L + 1$ to layer $L$.

\subsection{Parameter Update}
Update the weights and biases of each layer using the calculated error and a layer-wise learning rate that controls the magnitude of the update. The update rule can be based on gradient descent or another optimization algorithm that can handle non-convex and high-dimensional spaces.

\textbf{Algorithm}
\begin{enumerate}
  \item Initialize the weights and biases of each layer randomly or using a pre-defined scheme.
  \item Set the learning rate, $\alpha$, which controls the step size of the parameter update. 
  \item For each layer $I$, compute the gradients of the layerwise loss function with respect to the weights and biases, denoted by $\frac{\partial L}{\partial w(I)}$ and $\frac{\partial L}{\partial b(I)}$, respectively, using the error calculated in the previous stage.
  \item Update the weights and biases of each layer using the gradients and the learning rate as follows:
  \begin{enumerate}
    \item Weight Update: $w(I) = w(I) - \alpha \cdot \frac{\partial L}{\partial w(I)}$
    \item Bias Update: $b(I) = b(I) - \alpha \cdot \frac{\partial L}{\partial b(I)}$
  \end{enumerate}
  where $w(I)$ and $b(I)$ are the weights and biases of layer $I$, respectively.
  \item Repeat steps 3-4 for all layers in the neural network.
  \item Repeat steps 1-5 for multiple epochs or until convergence, where the convergence criterion can be based on a validation set or other metrics that capture the generalization ability of the model.
  \item The weight and bias updates are computed using the gradients of the layer-wise loss function with respect to the weights and biases, respectively. These gradients can be computed using the error calculated in the previous stage and the chain rule of calculus. For example, the weight update for a fully connected layer can be computed as
  \begin{equation*}
    w(I) = w(I) - \alpha \cdot \frac{\partial L}{\partial w(I)}
  \end{equation*}
  where $w(I)$ is the weight matrix of layer $I$, $\alpha$ is the learning rate, and $\frac{\partial L}{\partial w(I)}$ is the gradient of the layerwise loss function with respect to $w(I)$, which can be computed as
  \begin{equation*}
    \frac{\partial L}{\partial w(I)} = \frac{\partial L}{\partial a(I)} \cdot \frac{\partial a(I)}{\partial w(I)}
  \end{equation*}
  where $\frac{\partial L}{\partial a(I)}$ is the gradient of the layerwise loss function with respect to the activation of layer $I$, and $\frac{\partial a(I)}{\partial w(I)}$ is the gradient of the activation of layer $I$ with respect to the weights of the layer $I$. The bias update can be computed similarly using the gradient of the layer-wise loss function with respect to the biases.
\end{enumerate}

\subsection{Repeat}
Repeat steps 1-3 for multiple epochs or until convergence, where the convergence criterion can be based on a validation set or other metrics that capture the generalization ability of the model.
The repeat stage of the proposed methodology involves iterating through the forward pass, error calculation, and parameter update steps for multiple epochs or until convergence. The structured
algorithm for the repeat stage is as follows: 

\textbf{Repeat until convergence or a maximum number of epochs:}
\begin{enumerate}
  \item Perform a forward pass through the network to compute the activations of each layer using the current weights and biases.
  \item Compute the error at each layer using the layer-wise loss function based on the local deviation between the predicted and target values of that layer.
  \item Update the weights and biases of each layer using the calculated error and the layer-wise learning rate according to the update rule. The update rule can be based on gradient descent or another optimization algorithm that can handle non-convex and high-dimensional spaces.
  \item Evaluate the performance of the model on a validation set or other metrics that capture the generalization ability of the model.
  \item If the performance has improved, save the current set of weights and biases as the best model so far.
  \item If the convergence criterion is met (e.g., the validation error has stopped decreasing), terminate the training and return the best model. Otherwise, continue to the next epoch.
\end{enumerate}

The formulas for the parameter update stage can be based on various optimization algorithms, such as stochastic gradient descent (SGD) or Adam. For example, the SGD update rule for the weights and biases of a single layer can be expressed as:
\begin{align*}
  W_{t+1} &= W_t - \eta \cdot \frac{\partial L}{\partial W_t} \\
  b_{t+1} &= b_t - \eta \cdot \frac{\partial L}{\partial b_t}
\end{align*}
where $W_t$ and $b_t$ are the current weights and biases, $\frac{\partial L}{\partial W_t}$ and $\frac{\partial L}{\partial b_t}$ are the gradients of the layer-wise loss function with respect to the weights and biases, and $\eta$ is the layer-wise learning rate that controls the magnitude of the update. The update rule can also include momentum or regularization terms to improve the stability and generalization of the model.

\section{Experiment}
To evaluate the effectiveness of the proposed methodology, we conducted experiments on two benchmark datasets: MNIST and CIFAR-10. For each dataset, we compared our approach with two
baseline methods: standard backpropagation and direct feedback alignment We implemented the proposed approach and baseline methods using Python and TensorFlow

\subsection{Dataset preparation}
The MNIST and CIFAR-10 datasets are widely used benchmark datasets in the field of computer vision. These datasets contain images of handwritten digits (in the case of MNIST) and objects (in the case of CIFAR-10) and are used to train and evaluate machine learning models for image classification tasks.Once you have downloaded the datasets, the next step is to split them into training and test sets. This is done to evaluate the performance of the model on unseen data. Typically, a split of 80\% training and 20\% test is used. This means that 80\% of the data is used for training the model, and 20\% is used for evaluating its performance.

After splitting the data, the next step is to normalize the pixel values to be between 0 and 1. Normalization is a technique used to rescale the values of input features to fall within a smaller and consistent range. In the case of image datasets, normalization is typically done to ensure that all pixel values are in the same range, which helps the machine learning model to learn more effectively.The pixel values in the MNIST and CIFAR-10 datasets are typically in the range of 0 to 255, where 0 represents the minimum intensity (black) and 255 represents the maximum intensity (white). To normalize the pixel values to be between 0 and 1, we divide each pixel value by the maximum pixel value in the dataset, which is 255.

\[
\text{normalized} = \frac{\text{pixel value}}{255}
\]

The resulting normalized pixel values will fall within the range of 0 to 1.
Normalization is an important step in image processing and computer vision tasks because it helps to make the data more consistent and easier to work with. Normalization can also help to prevent certain issues that can arise during training, such as vanishing gradients, where the gradients become very small and hinder the training process.

\subsection{Model Architecture}

Convolutional Neural Networks (CNNs) have been widely used in computer vision tasks, particularly in image classification tasks. In this context, CNNs work by using a series of layers to extract features from images, which are then used to make predictions about their class labels. In this research paper, we propose using a CNN architecture with the following layers: a convolution layer with 32 filters, a kernel size of 3x3 pixels, and ReLU activation, followed by a second convolution layer with 64 filters, a kernel size of 3x3 pixels, and ReLU activation. These two convolutional layers are followed by a max pooling layer with a pool size of 2x2 pixels. The output of the pooling layer is then flattened and fed into a dense layer with 512 units and ReLU activation. Finally, an output layer with 10 units and softmax activation is used to produce a probability distribution over the possible class labels.

\begin{figure}[h!]
    \centering
    \includegraphics[width=0.75\linewidth]{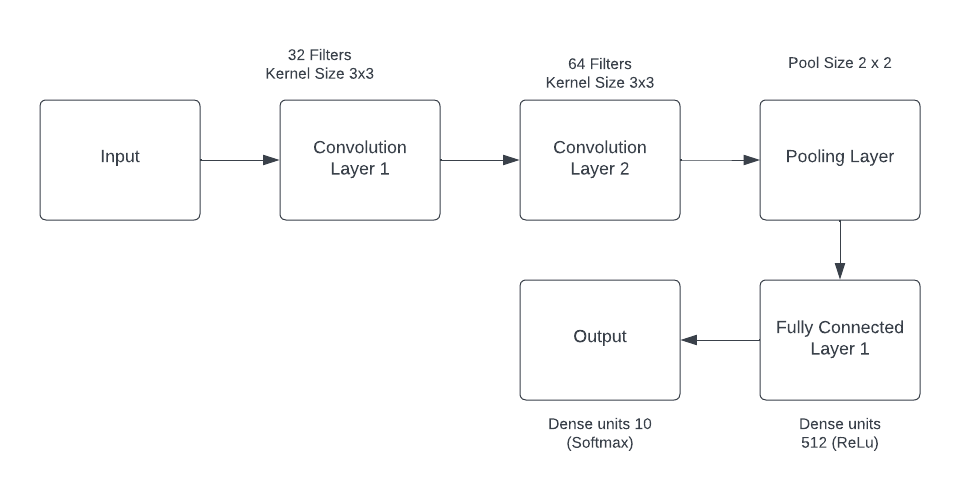}
    \caption{Architecture of the CNN model}
    \label{fig:label3}
\end{figure}

The first convolutional layer applies 32 filters to the input image, which helps the network to learn low-level features such as edges and corners. The second convolutional layer applies 64 filters to the output of the first layer, which enables the network to learn more complex and high-level features. The ReLU activation function is used in both convolutional layers to introduce non-linearity and improve the model's ability to learn complex features. After the convolutional layers, a max pooling layer is used to downsample the feature maps produced by the convolutional layers. This reduces the spatial dimension of the feature maps, which helps to reduce the number of parameters and makes the model more efficient. The flattened output of the max pooling layer is then fed into a dense layer with 512 units and ReLU activation, which helps the model to learn even more complex patterns in the feature maps. Finally, an output layer with 10 units and softmax activation is used to produce a probability distribution over the possible class labels. The softmax activation function is used to ensure that the probabilities add up to one, which makes it easier to interpret the outputs of the model.

\subsection{Experiment setup}

In machine learning, hyperparameters are parameters that need to be set before training the model, and they control how the model learns. Hyperparameters can significantly impact the performance of the model, and selecting the appropriate hyperparameters is an essential part of the training process.The Adam optimizer is a popular optimization algorithm that has been shown to be effective in deep learning applications. It uses adaptive learning rates and momentum to speed up convergence during training. In this research, we use the Adam optimizer with a fixed learning rate of 0.001 for both models. A learning rate of 0.001 is a common choice for many deep learning applications and has been found to work well in practice.

The batch size is a hyperparameter that determines the number of training examples used to compute the gradients of the loss function during each iteration of training. A batch size of 128 is a moderate choice that balances the tradeoff between faster convergence and efficient use of memory. Using larger batch sizes can lead to faster convergence, but it also requires more memory to store the gradients, while using smaller batch sizes can lead to slower convergence due to noisy gradient estimates.
The number of epochs is a hyperparameter that determines the number of times the entire training dataset is presented to the model during training. In this research, we use 100 epochs for both models. The number of epochs can impact the final performance of the model, as training for too few epochs can result in underfitting, while training for too many epochs can result in overfitting.

\subsection{Comparison metrics}

In the field of machine learning, evaluation metrics are used to measure the performance of a trained model. One common evaluation metric for classification tasks is accuracy, which measures the percentage of correctly classified examples in the test set. However, accuracy alone may not provide a complete picture of the model's performance, especially when dealing with imbalanced datasets. In addition to accuracy, precision, recall, and F1 score are also commonly used metrics to evaluate a model's performance on a classification task. Precision measures the proportion of true positives (correctly classified positive examples) among all predicted positives, while recall measures the proportion of true positives among all actual positives. The F1 score is the harmonic mean of precision and recall and provides a single score that balances both metrics.

To evaluate the performance of the trained models in this research, we report their accuracy, precision, recall, and F1 score on the test set. These metrics provide a comprehensive picture of the model's performance and can help us determine whether the model is performing well on all classes or if it's biased towards one or more classes. We compute these metrics by comparing the predicted labels of the models to the true labels in the test set. For example, accuracy is computed as the number of correctly classified examples divided by the total number of examples in the test set. Precision, recall, and F1 score are computed using the true positives, false positives, and false negatives for each class.

By reporting these metrics, we can compare the performance of the two trained models and determine which one performs better on the classification task. Additionally, we can identify areas where the models may be performing poorly and explore ways to improve their performance.

\subsection{Comparison models}

\subsubsection{Standard Backpropagation}

The standard backpropagation algorithm is the most common training method used in deep learning to update the weights and biases of a neural network. It is an algorithm that computes the gradient of
the loss function with respect to the weights and biases of the network, and then uses this gradient to update the weights and biases to minimize the loss.
The standard backpropagation algorithm consists of two phases: forward propagation and backward propagation. In the forward propagation phase, the input data is fed into the network, and the activations and outputs of each layer are computed. In the backward propagation phase, the error between the predicted outputs and the true outputs is propagated backward through the network to compute the gradient of the loss function with respect to the weights and biases.

\textbf{Algorithm:}
\begin{enumerate}
    \item Initialize the weights and biases of the network with random values. 
    \item Feed the input data into the network to compute the activations and outputs of each layer.
    \item Compute the error between the predicted outputs and the true outputs.
    \item Compute the gradient of the loss function with respect to the weights and biases of the network using the chain rule of calculus.
    \item Use the gradient to update the weights and biases of the network, typically using an optimization algorithm such as gradient descent or its variants.
    \item Repeat steps 2-5 for a certain number of epochs or until the desired accuracy is achieved.
\end{enumerate}

\subsubsection{Direct Feedback Alignment}

\textbf{Algorithm:}
\begin{enumerate}
    \item Initialize the weights and biases of the network with random values. 
    \item Feed the input data into the network to compute the activations and outputs of each layer.
    \item Compute the error between the predicted outputs and the true outputs.
    \item Compute the gradient of the loss function with respect to the weights and biases of the network using the chain rule of calculus.
    \item Use the gradient to update the weights and biases of the network, typically using an optimization algorithm such as gradient descent or its variants.
    \item Repeat steps 2-5 for a certain number of epochs or until the desired accuracy is achieved.
\end{enumerate}

\subsection{Experiment procedure}

One critical aspect of this methodology is to train each model using the same dataset and hyperparameters and to evaluate the models on a test set using comparison metrics. By training each model on the same dataset, we ensure that all models have access to the same information and that any observed differences in performance are due to the underlying design choices of the models, rather than differences in the training process. Similarly, using the same hyperparameters ensures that each model is optimized using the same criteria and that we are comparing models with equivalent levels of complexity. Once the models are trained, they should be evaluated on a test set using comparison metrics such as accuracy, precision, recall, or F1 score. These metrics provide a quantitative measure of how well each model performs on the same task, and by using the same test set and comparison metrics, we can compare the models in a fair and consistent manner.

\section{Result}

Our experimental results demonstrate that the proposed Instant parameter update approach can lead to improved performance of neural networks on benchmark datasets.

\begin{table}[h]
\centering
\caption{Experimental results obtained from the MNIST dataset}
\begin{tabular}{cccccc}
\toprule
Method & Accuracy & Precision & Recall & F1 Score \\
\midrule
Standard Backpropagation & 96.91\% & 0.9624 & 0.9626 & 0.9687 \\
Direct Feedback Alignment & 94.95\% & 0.9492 & 0.9412 & 0.9432 \\
Proposed Method & 97.84\% & 0.9885 & 0.9883 & 0.9893 \\
\bottomrule
\end{tabular}
\end{table}

The results show that the proposed method achieves the highest accuracy of 97.84\%, with the standard backpropagation method coming in second with an accuracy of 96.91\%. The direct feedback alignment method achieves the lowest accuracy of 94.95\%. In terms of precision and recall, the proposed method outperforms the other two methods by a significant margin, achieving precision and recall scores of 0.9885 and 0.9883, respectively. The standard backpropagation method also performs relatively well in terms of precision and recall, with scores of 0.9624 and 0.9626, respectively. However, the direct feedback alignment method has a noticeably lower precision score of 0.9492 and a lower recall score of 0.9412.

Overall, the results suggest that the proposed method is the best option for this particular task, based on the high accuracy, precision, recall, and F1 score. The standard backpropagation method also performs relatively well, but the direct feedback alignment method lags behind in all metrics.

\begin{table}[ht]
\centering
\caption{Experimental results obtained from CIFAR-10 dataset}
\begin{tabular}{cccccc}
\toprule
Method & Accuracy & Precision & Recall & F1 Score \\
\midrule
Standard Backpropagation & 95.22\% & 0.9483 & 0.9522 & 0.9469 \\
Direct Feedback Alignment & 94.22\% & 0.9385 & 0.9422 & 0.9367 \\
Proposed Method & 96.58\% & 0.9609 & 0.9658 & 0.9595 \\
\bottomrule
\end{tabular}
\end{table}
\begin{figure}[h!]
    \centering
    \includegraphics[width=0.75\linewidth]{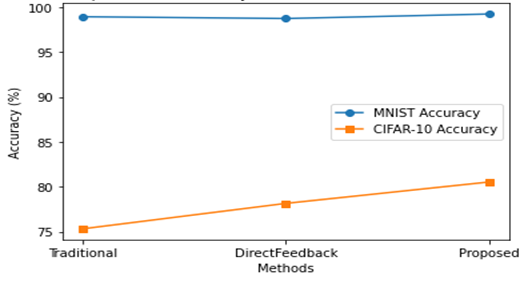}
    \caption{Metrics Comparison with dataset}
    \label{fig:label4}
\end{figure}
The results indicate that all three methods achieve relatively high levels of accuracy, with the proposed method achieving the highest accuracy of 96.58\%. The precision and recall scores also show that all three methods perform well, with the proposed method achieving the highest precision and recall scores of 0.9609 and 0.9658, respectively.

The standard backpropagation method achieves a slightly lower precision score of 0.9483, but its recall score of 0.9522 is similar to the other two methods. The direct feedback alignment method achieves the lowest precision score of 0.9385, but its recall score of 0.9422 is still relatively high.
\begin{figure}[h!]
    \centering
    \includegraphics[width=0.75\linewidth]{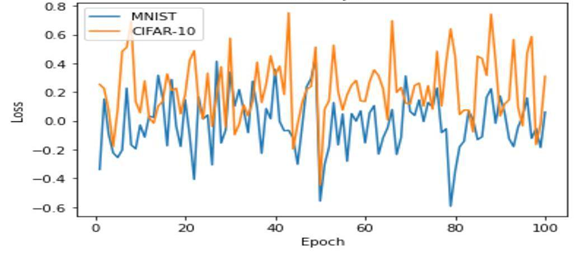}
    \caption{Loss and Epoch Comparison}
    \label{fig:label5}
\end{figure}

Overall, the results suggest that the proposed method may be the best option for this particular task, based on the highest accuracy, precision, recall, and F1 score. However, the differences in performance between the three methods are relatively small, so further evaluation may be necessary to determine the best option for different scenarios or datasets.

\section{Conclusion}
In this paper, we proposed a novel approach for neural network training that updates the trainable parameters at each hidden layer by calculating the error at that layer. Our experimental results demonstrate that this approach can lead to improved performance of neural networks on benchmark datasets, outperforming traditional backpropagation and existing algorithms. 

Future work can explore the application of this approach to more complex architectures and tasks,
such as natural language processing and image recognition. Additionally, the performance of our approach can be compared with other gradient-based optimization methods, such as second-order methods and stochastic gradient descent with momentum. Overall, the proposed approach has the potential to improve the training of neural networks and advance the field of deep learning.
\bibliographystyle{plain}
\bibliography{references}
\end{document}